\documentclass[conference]{IEEEtran}
\usepackage{times}
\usepackage{amsmath}
\usepackage{amsfonts}
\usepackage{graphicx}
\usepackage{caption} 
\usepackage[numbers]{natbib}
\usepackage{multirow,multicol}
\usepackage[bookmarks=true]{hyperref}
\begin{document}
\title{asRoBallet: Closing the Sim2Real Gap via Friction-Aware Reinforcement Learning for Underactuated Spherical Dynamics}
\author{\authorblockN{Fang Wan}
\authorblockA{Southern University of Science and Technology\\
Shenzhen, China\\
Email: wanfang@ieee.org}
\and
\authorblockN{Guangyi Huang, Tianyu Wu, Zishang Zhang, Bangchao Huang}
\authorblockA{Southern University of Science and Technology\\
Shenzhen, China}
\and
\authorblockN{Haoran Sun}
\authorblockA{The University of Hong Kong\\
Hong Kong SAR, China}
\and
\authorblockN{Mingdong Chen}
\authorblockA{AncoraSpring Robotics\\
Shenzhen, China\\
Email: chen@AncoraSpring.com}
\and
\authorblockN{Chaoyang Song}
\authorblockA{Mohamed bin Zayed University of Artificial Intelligence\\
Abu Dhabi, United Arab Emirates\\
Email: songcy@ieee.org}}


%
\maketitle
\begin{abstract}

    We introduce \textbf{asRoBallet}, to the best of our knowledge, the first end-to-end reinforcement learning (RL) locomotion policy deployed on a humanoid ballbot hardware platform. Historically, ballbots have served as a canonical benchmark for underactuated and nonholonomic control, which are characterized by a reality gap in complex friction models for wheel-ball-floor interactions. While current literature demonstrates successful handling of 3D balancing with LQR and MPC, transitioning to actual hardware for a humanoid ballbot using RL is currently hindered by critical gaps in contact modeling, actuator latency \& jitter, and safe hardware exploration. This study proposes a high-fidelity MuJoCo simulation that explicitly models the discrete roller mechanics of ETH-type omni-wheels, thereby capturing parasitic vibrations and contact discontinuities that have previously been ignored. We also developed a \emph{Friction-Aware Reinforcement Learning} framework that achieves \textbf{zero-shot Sim2Real transfer} by mastering the coupled rolling, lateral, and torsional friction channels at the wheel-ball and ball-floor interfaces. We designed asRoBallet through subtractive reconfiguration, repurposing key components from an overconstrained quadruped and integrating them into a newly designed structural frame to achieve a robust research platform at low cost. We also developed a generalized iOS ecosystem that transforms consumer electronics into a low-latency interface, enabling a single operator to orchestrate expressive humanoid maneuvers via intuitive natural motion.

\end{abstract}
\begin{IEEEkeywords}

    Humanoid BallBot, Tribology Modeling, Reinforcement Learning, Zero-Shot Sim2Real
    
\end{IEEEkeywords}
\IEEEpeerreviewmaketitle
\section{Introduction}
\label{sec:Intro}

    The single-ball-wheel robot, or ``ballbot,'' presents a theoretically optimal morphology for robotic assistants in human-centric environments \cite{Lauwers2006DynamicallyStable, Nagarajan2014TheBallbot}. Its omnidirectional mobility allows for agile navigation in constrained domestic spaces, while its tall, slender profile facilitates compliant physical interaction and social presence \cite{Song2023HandsFree, Xiao2023DesignControl}. However, its widespread deployment remains limited compared to quadrupedal or differential-drive systems due to the \textit{prohibitive complexity of custom hardware integration} and the \textit{Sim2Real gap inherent in controlling underactuated spherical dynamics}.

\subsection{Motivation and Related Work}

    The \textit{first} barrier lies in hardware architecture. Traditional ballbot platforms typically rely on expensive, custom-built drivetrains utilizing high-gear-ratio motors to ensure precise kinematic constraints \cite{Skrabel2013Rezero, Jespersen2019Kugle, Shu2021MomentumBased}. While effective at minimizing control uncertainty, this approach yields rigid, mechanically complex systems that lack impact resilience. Furthermore, perception often relies on dedicated industrial-grade LiDARs and robotic vision, driving costs beyond the reach of many research labs. This stands in stark contrast to recent advances in quadrupedal robotics, which have converged on proprioceptive quasi-direct-drive actuators \cite{Kau2019StanfordDoggo} and smartphone-based computing \cite{Oros2013Smartphone}. These architectures offer high torque density, back-drivability, and significant cost reductions. However, the application of such compliant actuation ecosystems to the inherently unstable dynamics of a ballbot remains underexplored.

    The \textit{second}, and perhaps more critical, barrier is the \textit{Reality Gap} in control. Classical control approaches for ballbots, such as Linear Quadratic Regulators (LQR) \cite{Nagarajan2014TheBallbot, Jespersen2019Kugle} or Model Predictive Control (MPC) \cite{Jespersen2020PathFollowing, Studt2022ParameterIdentification}, rely on linearized dynamic models (e.g., the Planar Inverted Pendulum). These controllers operate under strict assumptions: infinite friction at the contact point (no slip), negligible rolling resistance, and ideal actuation. In reality, the interface between a rubber-coated ball and omni-directional rollers is governed by complex tribology, specifically the chaotic coupling of rolling resistance, lateral slip, and torsional friction \cite{Bhushan2013Introduction}. 

    When coupled with the passive dynamics of the rollers and the floating ball, the system explodes in complexity. A single ETH-type omni-wheel with multiple rows of rollers could introduce 10 to 12 passive Degrees of Freedom (DoF). Standard model-based controllers struggle to account for these microdynamics. Reinforcement learning has enabled end-to-end sim-to-real locomotion policies, particularly for quadruped and biped robots. However, RL-based ballbot locomotion remains underexplored, with prior studies limited to restricted control settings rather than deployable locomotion policies. \cite{farshidian2014learning} used RL to refine an iLQG-based controller on the Rezero ballbot, demonstrating hardware adaptation but relying on a model-based initialization and optimizing a finite-horizon motion controller. \cite{Zhou2021LearningBallBalancing} proposed a compound controller in which a DDPG policy augments a conventional feedback controller to improve balance recovery, but the method was evaluated only in simulation and remained limited to the recovery task. \cite{buzzetti2024reinforcement} et al. trained a PPO policy for fall-damage reduction on a simulated ballbot, where the policy controlled only the arms during falling. More recently, \cite{Salehi2025ReinforcementLearning} introduced a MuJoCo-based ballbot simulation by modeling omniwheels as capsules with anisotropic tangential friction and demonstrated PPO-based navigation only in simulation as a proof of concept, leaving sim-to-real transfer to future work.
    
    \textit{Finally}, a significant usability gap exists in the teleoperation of such high-Degree-of-Freedom systems, which typically require complex, non-intuitive interfaces (e.g., joystick arrays) and often necessitate multiple operators to coordinate locomotion and upper-body gestures separately. There is a distinct lack of lightweight interfaces that leverage ubiquitous consumer hardware to provide unified, whole-body control.

\subsection{Key Contributions}

    To bridge these gaps in hardware accessibility, control robustness, and interface usability, this paper presents asRoBallet, a holistic ballbot system for robust humanoid balancing. Our specific contributions are as follows:
    \begin{enumerate}
        \item We present a cost-effective, high-bandwidth humanoid ballbot design achieved by reconfiguring the proprioceptive actuation, computation core, and mobile perception of an overconstrained quadruped into an underactuated spherical morphology.
        \item We introduce granular dynamics that explicitly simulate the discrete roller mechanics of ETH-type omni-wheels, successfully capturing the parasitic vibrations and roller handovers critical to physical balancing stability.
        \item We propose a friction-aware reinforcement learning framework that achieves zero-shot hardware deployment by mastering the tribological couplings of rolling, lateral, and torsional friction at the wheel-ball-floor interfaces, enabling robust whole-body locomotion.
    \end{enumerate}

    We also release an iOS app (asMagic: \url{https://apps.apple.com/us/app/asmagic/id6661033548}) that leverages consumer-grade sensing for low-latency, markerless whole-body teleoperation, facilitating intuitive human-robot interaction across diverse humanoid platforms. Public release of the asRoBallet will be updated here (\url{https://github.com/asRoBallet/asRoBallet_mujoco}).

\section{Reconfigurable Design of asRoBallet}
\label{sec:Design}

\begin{figure}[htbp]
        \centering
        \includegraphics[width=1\linewidth]{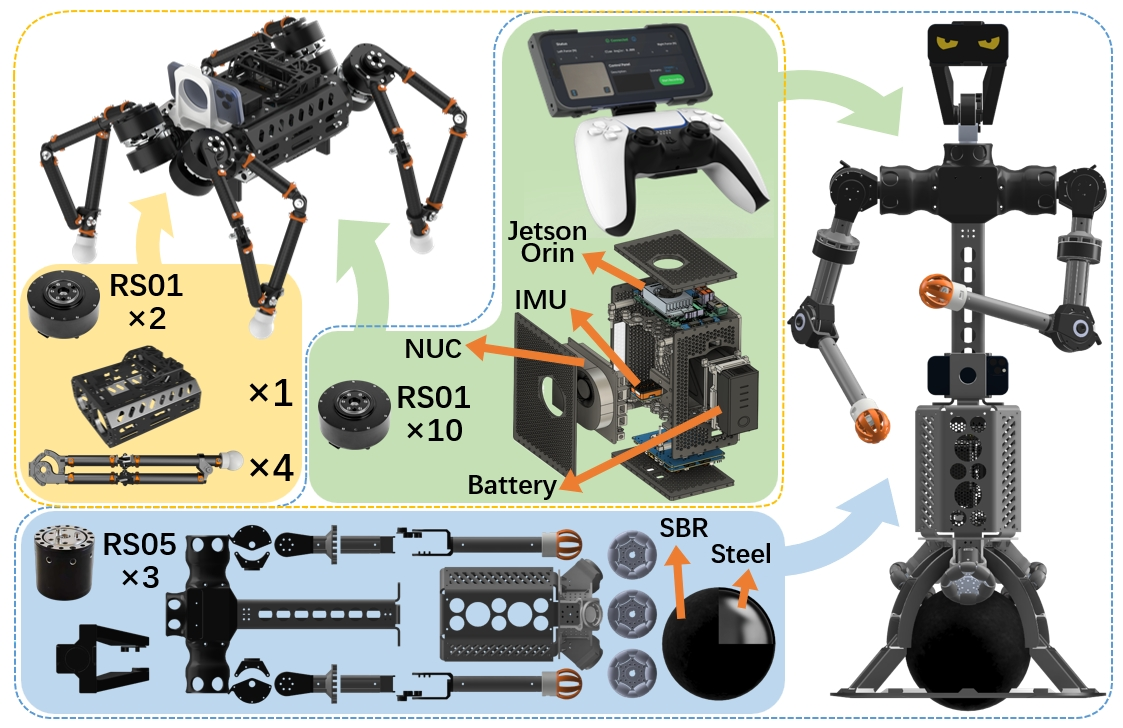}
        \caption{
            Structural transformation from the asOverDog quadruped (top left) to the asRoBallet humanoid (right).
        }
        \label{fig:Design_asRoBallet}
    \end{figure}

    We employ a \textit{subtractive reconfiguration} methodology to construct asRoBallet, transforming the open-source \textit{asOverDog} quadruped \cite{Sun2024OverconstrainedLocomotion, Gu2025OneDoF} into a humanoid ballbot. By retaining the high-bandwidth actuation, computation, and perceptual core while redesigning the structural chassis, we achieve a cost-effective, impact-resilient platform, as shown in Fig. \ref{fig:Design_asRoBallet}.
    
\subsection{Mechanical Architecture: From Quadruped to Ball}
    
    The design transformation primarily replaces the chassis and four overconstrained limbs to build asRoBallet.
    
\subsubsection{Spherical Wheel Base}
    
    Three ETH-type omni-wheels (12 rollers each, offset by $120^{\circ}$) drive a 220 mm SBR-coated hollow steel ball. To minimize starting torque and cogging, we replaced the original quadruped drive units with lower-torque actuators (RobStride 05) rated at 5.5 Nm.
    
\subsubsection{Humanoid Upper Body \& Frame}
    
    The remaining high-torque actuators (RobStride 01, 17 Nm) are reconfigured to form a 2-DoF head and two 4-DoF arms with tactile spheres capable of Vision-based Deformable Perception (VBDeformP) \cite{Wan2025SeeThruFinger, Han2025AnchoringMorphological}. These are mounted on a lightweight T-shaped sheet-metal frame with a 3D printed shoulder designed to minimize rotational inertia relative to the Center of Mass (CoM). The final system comprises 49 degrees of freedom (13 active, 36 passive), presenting a highly underactuated control challenge.
    
\subsection{Proprioceptive \& Computational Core}
    
    We transplant the electro-mechanical stack directly to accelerate commissioning. The quasi-direct-drive actuators provide intrinsic proprioception via current-based torque estimation, which is essential for our friction-aware policy. The self-contained computational unit houses a DJI TB47S battery, a 2 kHz industrial IMU (XSense MTI630), an Intel NUC for real-time control, and a Jetson Orin for neural inference, eliminating the need for custom power electronics.
    
\subsection{Consumer-Grade Perception \& Interaction}
    
    To minimize cost and complexity, we replace industrial sensing arrays with a unified consumer-grade ecosystem.
    
\subsubsection{Visual-Inertial Odometry (VIO)}
    
\begin{figure}[htbp]
        \centering
        \includegraphics[width=1\linewidth]{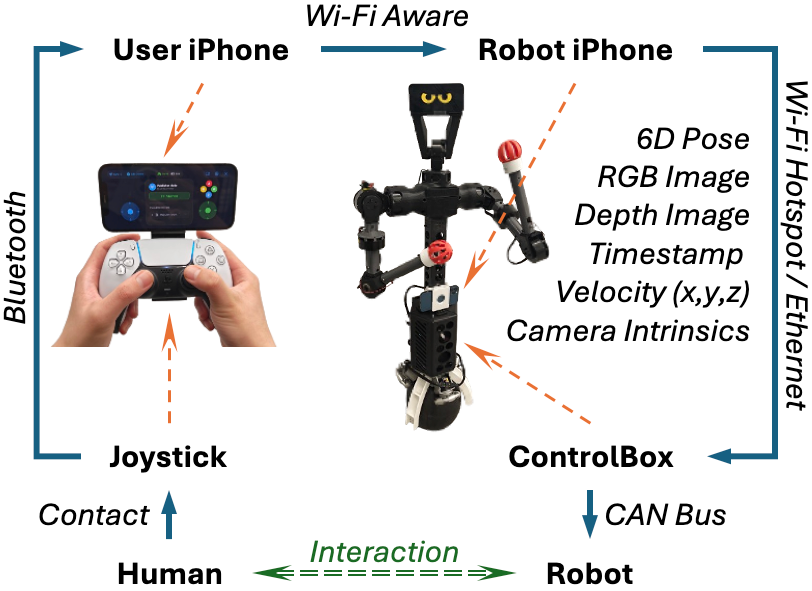}
        \caption{
            Repurpose mobile perception \& computing from an iPhone for robotics, the same for asRoBallet \& asOverDog.
        }
        \label{fig:Design_iOS}
    \end{figure}

    An iPhone 12 Pro mounted on the spine (Fig. \ref{fig:Design_iOS}) serves as the primary state estimator. By offloading VIO to the device's Neural Engine, we achieve low-drift 6-DoF localization that feeds directly into the controller, eliminating the need for discrete LiDARs.

\subsubsection{Scalable Teleoperation}
    
    We developed a custom iOS application that uses Wi-Fi Aware for low-latency peer-to-peer communication. The interface ranges from simple gamepad control to embodied telepresence, in which a distributed network of Apple devices (Watch, AirPods, iPhone) captures markerless human motion, enabling future generalization for intuitive whole-body retargeting.

\section{Friction-Aware Modeling and Reinforcement Learning of Underactuated Spherical Dynamics}
\label{sec:Learning}

\subsection{High-Fidelity Modeling in MuJoCo}
\label{sec:Learning_Sim}

\begin{figure}[htbp] 
        \centering
        \includegraphics[width=1\linewidth]{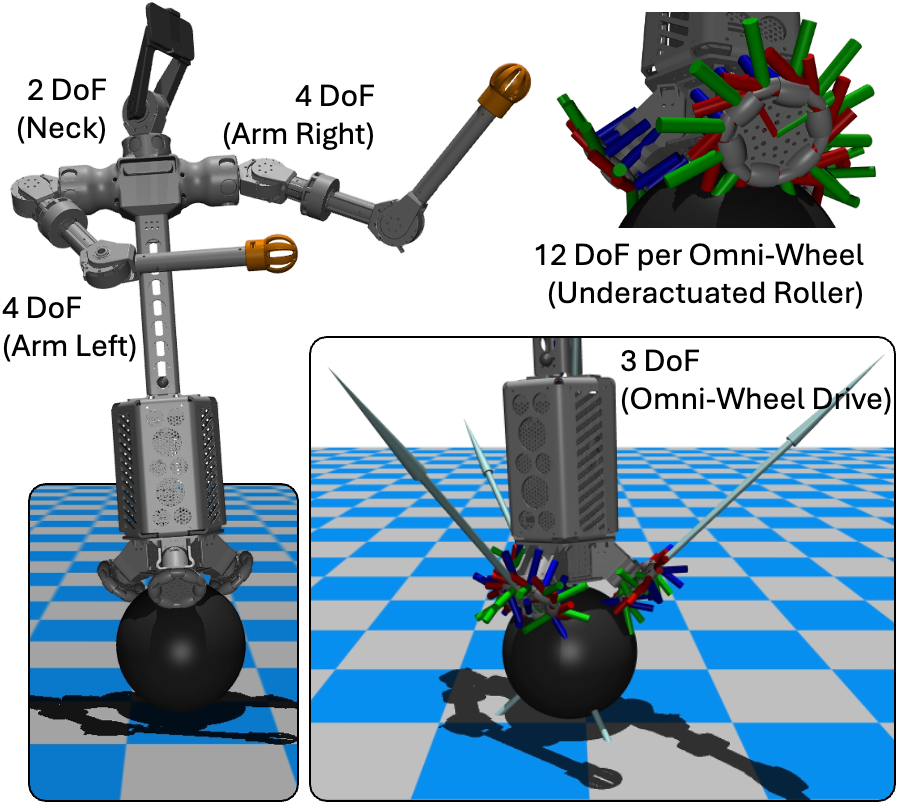}
        \caption{
            High-fidelity MuJoCo simulation of asRoBallet's highly underactuated spherical dynamics.
        }
        \label{fig:Learning_Sim_MuJoCo}
    \end{figure}

    Analytical models typically assume ideal rolling constraints (no slip) and negligible rolling resistance, thereby ignoring complex tribological phenomena, including the hysteresis of the rubber coating, the anisotropic slip of the omni-rollers, and the nonlinear friction of the actuators. To bridge the reality gap, we construct a high-fidelity MuJoCo multibody simulation (see Fig. \ref{fig:Learning_Sim_MuJoCo}) that models these uncertainties.

    We model asRoBallet with a floating base comprising the upper torso, three omni-wheels, and a spherical wheel. For training, we added a virtual, low-damped ball joint between the torso and the ball to ensure stable contact during action exploration. The ball joint is set to ensure that the total contact force between wheels and ball equals the torso's weight, thereby reproducing the real robot's contact state. We experimentally found that training fails without the ball joint. Once the omni-wheels are no longer in contact with the ball, it is almost impossible to further improve the model or the reward. We will later demonstrate that the policy trained with this virtual ball joint is sufficiently robust to be directly transferred to simulation and real-world hardware tests.
    
    \textbf{Omni-wheel Modeling.} We explicitly model omni-wheels' physical geometry. Each wheel is constructed as a ring of 12 free-spinning rollers (alternating between load-bearing and spacer rollers) attached to the main hub via hinge joints. This geometric fidelity is crucial for capturing \textit{lateral compliance}. Holonomic motion in the asRoBallet hardware is achieved by allowing the rollers to spin freely orthogonal to the drive direction. In our simulation, this is not a mathematical abstraction but a physical outcome of the physics of roller–ball contact. We employ MuJoCo's soft-constraint convex solver with an elliptic friction cone (\texttt{cone=``elliptic''}) and a high constraint impedance ratio (\texttt{impratio=10}) to resolve the overconstrained nature of the three-point drive system without numerical instability. To handle the stick-slip transitions inherent to balancing, we enable the Newton solver with \texttt{noslip\_iterations=1}.

    \textbf{Solver \& Contact Formulation.} We carefully model frictions at two critical contact interfaces. For the ball–floor contact pair, we specify full six-dimensional contact (normal, two tangential directions, torsional spin, and two rolling resistances) with relatively high sliding $\mu_{slide}=1$ and spin $\mu_{torsion}=0.01$ (interpreted as the effective diameter of the contact patch) friction but small rolling resistance $\mu_{roll}=0.0001$. This choice prevents the ball from spinning freely in place, yields a realistic rolling onset, and helps the robot ``bite'' the floor when generating yaw torques. At the wheel-ball interface, we use standard three-dimensional contact (normal plus two tangential directions). We model a soft \textit{rubber-like} layer on the ball by making contacts start stiff (impedance 0.85) and become very stiff (0.99) by 3 mm penetration (\texttt{solimp=``0.85 0.99 0.003''}). Lateral compliance is provided by the rollers' free joints rather than by anisotropic friction coefficients, thereby keeping the model both faithful and numerically robust. 

    \textbf{Actuator Frictions.} We further identify the friction parameters of the omni-wheel actuators under no-load, including the viscous friction coefficient $D_v$ and the Coulomb friction torque $D_c$, which is the minimum torque required to rotate the omni wheel. An increasing torque was given to the motor, and the motion of the motor can be described by \cite{Nagarajan2014TheBallbot}:
    \begin{equation}
        I_{motor}\ddot{\theta} = \tau(t)-D_v\dot{\theta}-D_c, \dot{\theta}>0.
    \end{equation}
    In the station-keeping task, the omni-wheels rotate slightly, with small motor torques typically oscillating around zero. Hence, it is especially sensitive to the Coulomb friction torque $D_c$. We experimentally determine $D_c$ in our system to be within the range of 80 to 120 mNm, which is much higher than those reported for Rezero (42 mNm) \cite{Doessegger2010Rezero} and Kugle (26 mNm) \cite{Jespersen2019Kugle}, which use much more expensive motors. The relatively large breakaway torque creates a dead zone in our wheel torques. Tiny corrective torques from the controller/policy do not move the wheels until the error grows enough to exceed that threshold, then the wheel \textit{breaks free}, producing a bigger-than-intended correction. This manifests as position limit-cycle jitter. 

    We add a body frame ${B}$ at the ball center, but rigidly attached to the body, providing linear velocity in local coordinates. We also add an IMU frame to host the gyro and orientation sensors, and an iPhone frame providing localization in world coordinates ${W}$. Another interesting finding is the selection of actuation types. Following previous work on simulating robotic arms and quadrupeds, the joints in the two arms and neck are actuated using position servos \cite{Zakka2025MuJoCoPlayground}. However, the three omni-wheels are actuated using direct motor torques. We experimentally find that training fails when a motor-type actuator is replaced with a velocity-type actuator. 

\subsection{Reinforcement Learning}
\label{sec:Learning_RL}

    \textbf{Task Formulation.} We define two distinct locomotive tasks: \textit{Velocity Tracking} and \textit{Station Keeping} for long-term hovering. We decouple the upper-body manipulation from the base stabilization to simplify the learning landscape. During locomotion training, the arms and head are treated as passive inertial payloads subject to randomized configurations. This makes the base policy robust to significant shifts in the Center of Mass (CoM) without requiring a unified whole-body policy, thereby reducing the dimensionality of the learning problem.

    \textbf{State and Action Space.} The action space $\mathcal{A} \in \mathbb{R}^3$ are continuous torque commands for the three omni-wheel motors. The observation vector $o_t$ differs slightly between tasks.
    \begin{itemize}
        \item \textbf{Velocity Tracking ($o_t^{VT}\in\mathbb{R}^{16}$):} are defined with respect to the body coordinate, thus requiring only proprioceptive measurements.
        \begin{equation}
            o_t^{VT} = \big[ \mathbf{v}_{B}, \hat{\mathbf{v}}_{cmd}, \mathbf{q}_{tilt}, \mathbf{x}_{\mathrm{COM}}, \boldsymbol{\omega}_{B}, \mathbf{a}_{t-1} \big]^\top
        \end{equation}
        Here, $\mathbf{v}_{B} \in \mathbb{R}^2$ is the body frame's estimated linear velocity, $\hat{\mathbf{v}}_{cmd}$ is the commanded velocity $(\hat{v}_x,\,\hat{v}_y,\,\hat{\omega}_z)$, $\mathbf{q}_{tilt}$ is the roll and pitch tilting relative to gravity, $\boldsymbol{\omega}_{B}$ is the gyroscope angular rate, and $\mathbf{a}_{t-1}$ is the previous action history. $\mathbf{x}_{\mathrm{COM}}$ is Center of Mass (COM) in the body frame {B} such that its value only reflects the change of arm configurations, given by 
        \begin{align}
            ^B\mathbf{x}_{\mathrm{COM}} &= ^W_BR^T\left(^W\mathbf{x}_{\mathrm{COM}} - ^W\mathbf{x}_{ball} \right), \\
            ^W\mathbf{x}_{\mathrm{COM}} &= \frac{1}{M}\sum_{i=1}^{n_b} m_i\,^{W}\mathbf{x}_i, \\
            M&=\sum_{i=1}^{n_b} m_i 
        \end{align}
        where $^W_BR^T$ is the rotation matrix of the body with reference to the world frame ${W}$, $m_i$ and $^{W}\mathbf{x}_i$ are the mass and world-frame COM position of body $i$ in MuJoCo.
        
        \item \textbf{Station Keeping ($o_t^{SK}\in\mathbb{R}^{17}$):} We augment proprioceptive state with exteroceptive position data provided by the iPhone's VIO pipeline.
        \begin{equation}
            o_t^{SK} = \big[ \mathbf{v}_{B}, \mathbf{p}_{W}, \sin{\psi}, \cos{\psi}, \mathbf{q}_{tilt}, \boldsymbol{\omega}_{B}, \mathbf{a}_{t-1} \big]^\top
        \end{equation}
        where $\mathbf{p}_{W} \in \mathbb{R}^2$ is the planar position error in the world frame and $\psi$ is the yaw error.
    \end{itemize}
    
    A critical challenge in the observation design is estimating $\mathbf{v}_{B}$. In simulation, the ball's linear velocity in the body frame is measured by a velocimeter mounted at the body frame ${B}$. In real experiments, we approximate this via non-slip forward kinematics at the wheel-ball interface.

    \textbf{Reward Design.} As shown in Table \ref{tab:Learning_Reward}, the rewards for velocity tracking consist of (a) tracking reward of the target linear and angular velocities ($\sigma$=0.07), (b) penalty for roll/pitch angular velocities, (c) an uprightness penalty to prevent excessive leaning, and (d) a smoothness penalty on control inputs (e.g. $-\lambda |u_t - u_{t-1}|^2$ to discourage abrupt changes in action). The smoothness term directly promotes graceful motion by implicitly minimizing jerk. For station-keeping, the rewards include terms for (a) minimizing the distance of the robot from the goal position (current position when asRoBallet is asked to stay station), (b) minimizing the tilt angle (uprightness), and (c) low control effort and smoothness. 

    \begin{table}[htbp]
        \caption{Summary of reward terms.}
        \label{tab:Learning_Reward}
        \begin{tabular}{llll}
        \hline
        Task & Term & Weight & Expression \\ \hline
    
        \multirow{6}{*}{\begin{tabular}[c]{@{}l@{}}Velocity \\ Tracking\end{tabular}}
            & Linear Vel.   & 0.5  & $\exp(-\left\| \boldsymbol{v_{xy}}-\boldsymbol{\hat{v}_{xy}} \right\|^2/{\sigma_{xy}}^2)$ \\
            & Yaw Ang. Vel. & 0.5  & $\exp(-\left\| \omega_{z}-\hat{\omega}_{z} \right\|^2/{\sigma_{w}}^2)$  \\ 
            & Angular Vel.  & -0.1 & $\left\|\boldsymbol{\omega_{xy}} \right\|^2$ \\ 
            & Torque        & -0.1 & $\left\|\boldsymbol{\tau} \right\|^2$ \\ 
            & Action Rate   & -0.1 & $\left\|\boldsymbol{a_{t}}-\boldsymbol{a_{t-1}} \right\|^2$ \\
            & Termination   & 1    & $\mathbb{1}(\phi,\theta \leq 20^{\circ})$ \\ \hline
    
        \multirow{6}{*}{\begin{tabular}[c]{@{}l@{}}Station\\ Keeping\end{tabular}}
            & Pos. \& Heading & 1      & $1-\tanh(\left\| (x,y,\psi) \right\|)$ \\
            & Angular Vel.    & -0.001 & $\left\|\boldsymbol{\omega} \right\|^2$ \\
            & Body Tilt       & -0.001 & $\left\| (\phi,\theta) \right\|^2$ \\
            & Torque          & -0.01  & $\left\|\boldsymbol{\tau} \right\|^2$ \\
            & Action Rate     & -0.01  & $\left\|\boldsymbol{a_{t}}-\boldsymbol{a_{t-1}} \right\|^2$ \\
            & Termination     & 1      & $\mathbb{1}(\phi,\theta \leq 20^{\circ})$ \\ \hline
    
        \end{tabular}
    \end{table}

    \textbf{Episode Reset \& Termination.} For velocity tracking, episode-wise target commands are sampled by a masked update rule. A fresh proposal $y_k\!\sim\!\mathcal{U}(V_{min},V_{max})^3$ is filtered by two independent Bernoulli masks $z_k,w_k\sim\text{Bernoulli}(0.5)$ per dimension, yielding
    \begin{equation}
        x_{k+1} = x_k - w_k\odot(x_k - z_k\odot y_k),
    \end{equation}
    which either holds, zeros, or replaces each component with a new value drawn from the distribution. This command is injected at reset and remains constant during the episode. The pose and velocity of the body are reset at every episode, following: 
    \begin{enumerate}
        \item yaw: $\psi\sim\mathcal{U}(-\pi,\pi)$, 
        \item roll/pitch: $\theta_x,\theta_y\sim\!\mathcal{U}(-5,5)$ degree, 
        \item horizontal linear velocities: $v_x,v_y\sim\mathcal{U}(-0.5,0.5)$ m/s, 
        \item angular velocities: $w_x,w_y,w_z\sim\mathcal{U}(-0.1,0.1)$ rad/s, 
        \item head/arms configurations uniformly sampled from the control range of each joint. 
    \end{enumerate}
    Episodes terminate early if roll or pitch exceeds $20^\circ$; otherwise, they truncate at a horizon of $1,000$ steps (equal to 10 seconds). The arm collision is ignored during configuration randomization, since we do not use the joint values; instead, we randomize the COM dynamics.
    
    For station-keeping tasks, the targeted position and heading direction are set to zero. The reset is similar to velocity tracking except the first two items are replaced by resetting the initial pose in yaw of $\psi\sim\mathcal{U}(-\pi/6,\pi/6)$ and location of $x,y\sim\mathcal{U}(-0.5,0.5)$m.  

    \textbf{Domain Randomization \& Robustness.} Previous work has shown that ballbots are highly sensitive to noises in the body's angular velocities, resulting in tremors during hardware deployment, thereby requiring post-processing filters, such as low-pass and Kalman filters. Classic model-based controllers for ballbots, such as those at CMU and Kugle, are typically designed under the no-slip (high-friction) assumption and linearize the dynamics around that constraint to enable clean LQR/MPC designs with precise torque tracking. When friction drops (due to dust, smooth tile, or moisture), their assumptions may break. We implement a randomization routine that samples (a) the ball-floor tangential friction coefficient from $\mathcal{U}(0.6,1.2)$, and (b) the motor's Coulomb friction torque from $\mathcal{U}(0.08,0.12)$ Nm, (c) gyroscope sensor noises from $N(0, 0.009)$ rad/s, matching real IMU's noise level. 
    
    \textbf{Learning Algorithm \& Training Setup.} All policies were trained with PPO using Stable-Baselines3. We used 8 parallel MuJoCo environments for data collection, with 2048 rollout steps per environment, yielding 16,384 samples per PPO update. The policy was optimized with a mini-batch size of 512 for 10 epochs per update. Each environment step corresponds to \texttt{frame\_skip}=5 MuJoCo simulation steps with a simulator time step of 0.002 s, resulting in a 100 Hz control frequency. The velocity-tracking task used \texttt{max\_episode\_steps = 1000}, corresponding to a maximum episode duration of 10 s, while the station-keeping task doubles the episode length. The longer horizon for station keeping encourages sustained balance and drift suppression, whereas the shorter horizon for velocity tracking increases the diversity of sampled command-following trials during training. Both policies were trained for $4.0\times10^6$ environment steps. Training was performed on a workstation with an Intel Core i9-13900KF CPU; the wall-clock training time was approximately 2.5 hours.

    The actor and critic use separate multilayer perceptrons with the same architecture: two hidden layers of 64 units with Tanh activations, followed by a linear output layer. The actor outputs three continuous omni-wheel torque commands, clipped to $[-1,1]$.

\section{Humanoid Interaction with an iOS App}
\label{sec:iOSApp}

    This section introduces an iOS application we developed for asRoBallet, asMagic, providing (1) an All-in-One real-time sensory suite, (2) a scalable interface for humanoid interaction, and (3) a general-purpose visualization for humanoids. 

\subsection{iPhone as an All-in-One Real-time Sensing Suite}
    
    Robots require rich real-time sensory feedback for physical intelligence, but building such perceptual capability remains costly and difficult to reuse across embodiments \cite{Bonci2021HumanRobot, Wang2024CrossEmbodiment, Doshi2024ScalingCross}. Recent advances in mobile computing offer a low-cost alternative by integrating diverse sensors into consumer smartphones. In particular, the iPhone provides mature embedded sensing, software APIs, and deployment infrastructure. In this work, we present an iOS app that explores this opportunity, leveraging mobile perception as an \textit{All-in-One} sensing solution for advanced robotics. 

    \begin{figure}[htbp]
        \centering
        \includegraphics[width=1\linewidth]{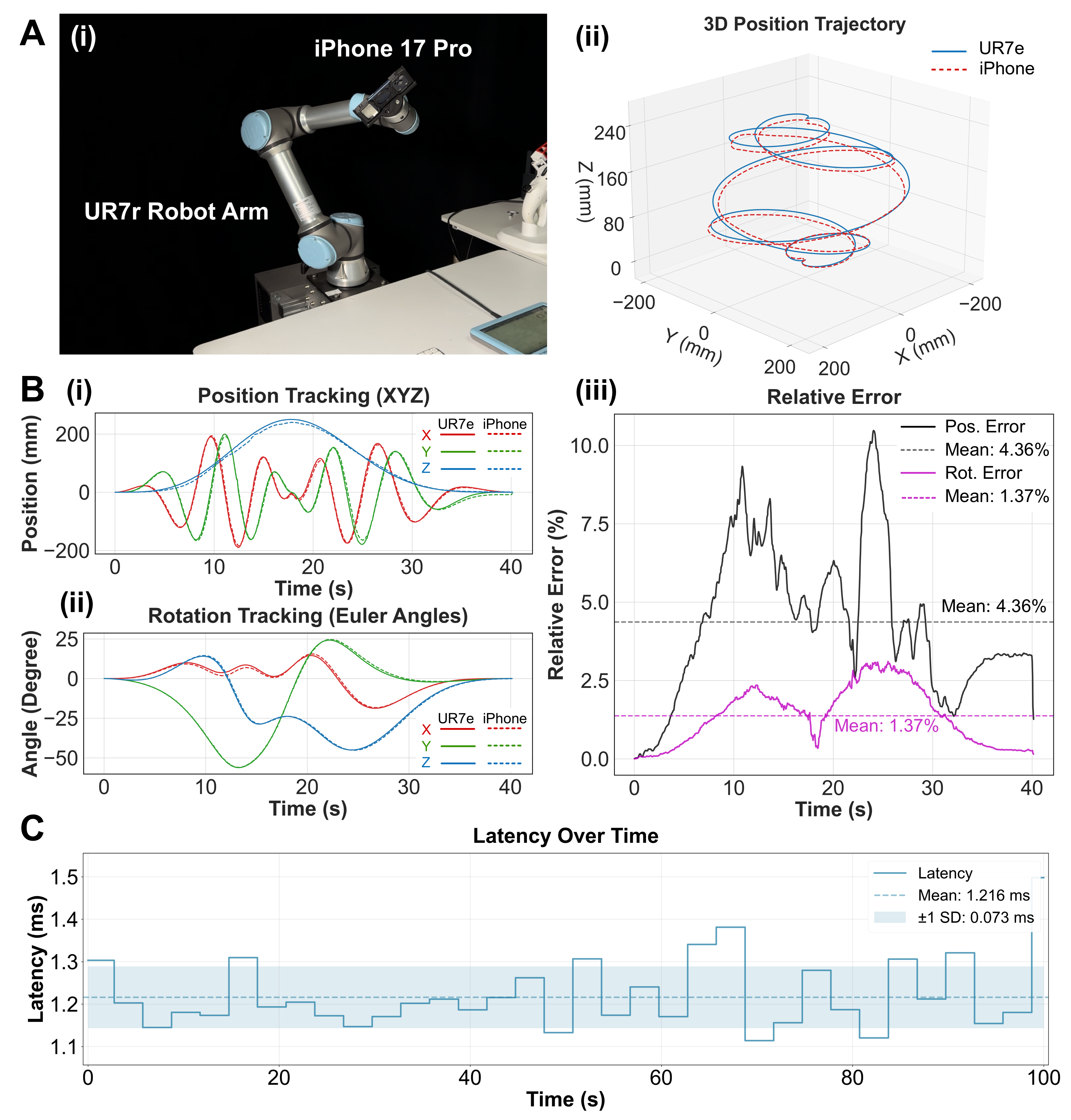}
        \caption{
            iPhone's ARKit pose estimation performance and peer-to-peer latency results.
        }
        \label{fig:iOSApp_LidarLatency}
    \end{figure}

    The app provides extensive APIs to access sensors on Apple devices, including iPhone, Apple Watch, iPad, AirPods, and Mac. We evaluate of iPhone 17 Pro's pose estimation performance against ground truth collected from an UR7e robotic arm (Fig. \ref{fig:iOSApp_LidarLatency}A). We observed promising relative errors in position tracking (within 5\%) and orientation tracking (within 1.5\%) in Fig. \ref{fig:iOSApp_LidarLatency}B, both of which are sufficiently competitive to consider the iPhone Pro as a high-performing sensor for advanced robotics. We also tested latency performance between the iPhone 16 Pro Max and the iPhone 17 Pro. The average latency was approximately 1.2 ms under common usage conditions, measured over a 100 s communication session at 100 Hz, as shown in Fig. \ref{fig:iOSApp_LidarLatency}C. This performance is consistent over an extended period of use but drops significantly with heating. Adding active cooling resolves this issue.

\subsection{A Scalable Interface for Humanoid Interaction}

    The anthropomorphic design of humanoids poses interaction challenges due to the lack of an interface to retarget expressive human movements to the humanoid's multi-DoF execution. A recent report from Disney Research indicates that 4 operators, each using a different interface, are required to achieve expressive, lively interaction with the H.E.R.B.I.E. humanoid in filmmaking. The majority of commercial humanoids are controlled using joystick-like devices for preprogrammed actions, which requires additional support for whole-body interaction. 

\begin{figure}[htbp]
        \centering
        \includegraphics[width=1\linewidth]{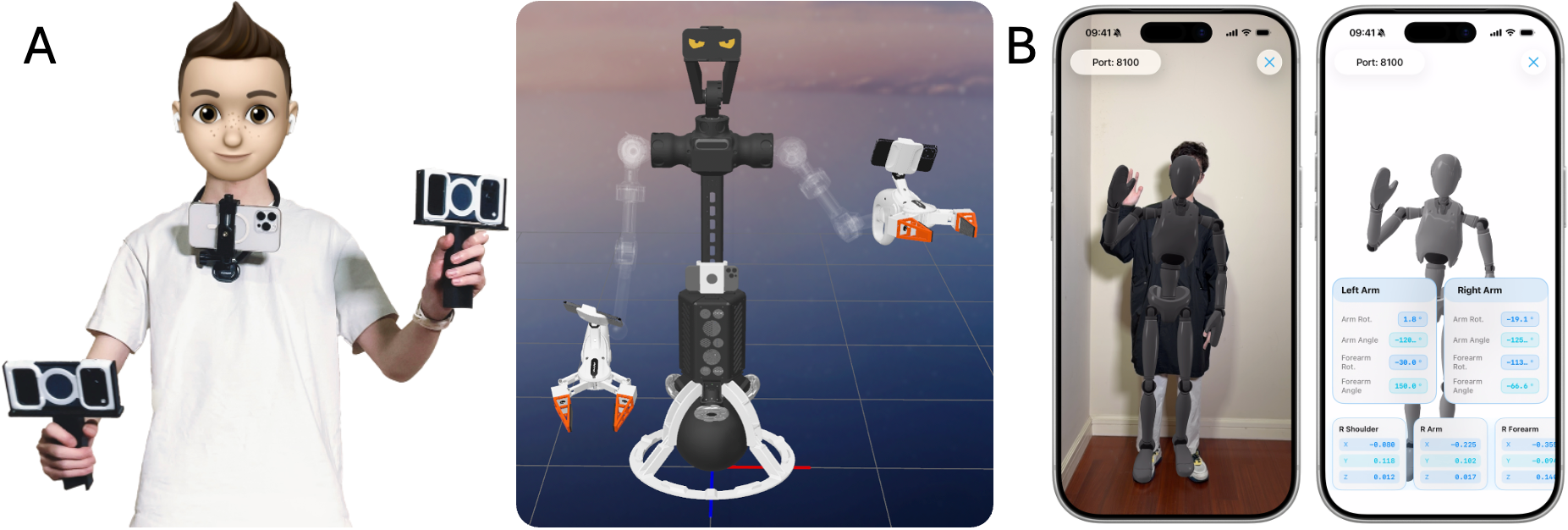}
        \caption{
            Scalable Human-Robot Interaction via iOS \& our app.
        }
        \label{fig:iOSApp_Interface}
    \end{figure}

    We propose a scalable iOS-based interface that enables three levels of direct humanoid interaction: (1) simple control as a joystick (Touchscreen control in Fig. \ref{fig:Design_iOS} without using a physical joystick), (2) wearable motion capture as an IMU suit (Fig. \ref{fig:iOSApp_Interface}A), and (3) markerless motion capture via whole-body pose tracking (Fig. \ref{fig:iOSApp_Interface}B), all achieved by a single operator using iOS devices.

\subsection{A General Purpose Visualization for Humanoid Robots}

\begin{figure}[t]
        \centering
        \includegraphics[width=1\linewidth]{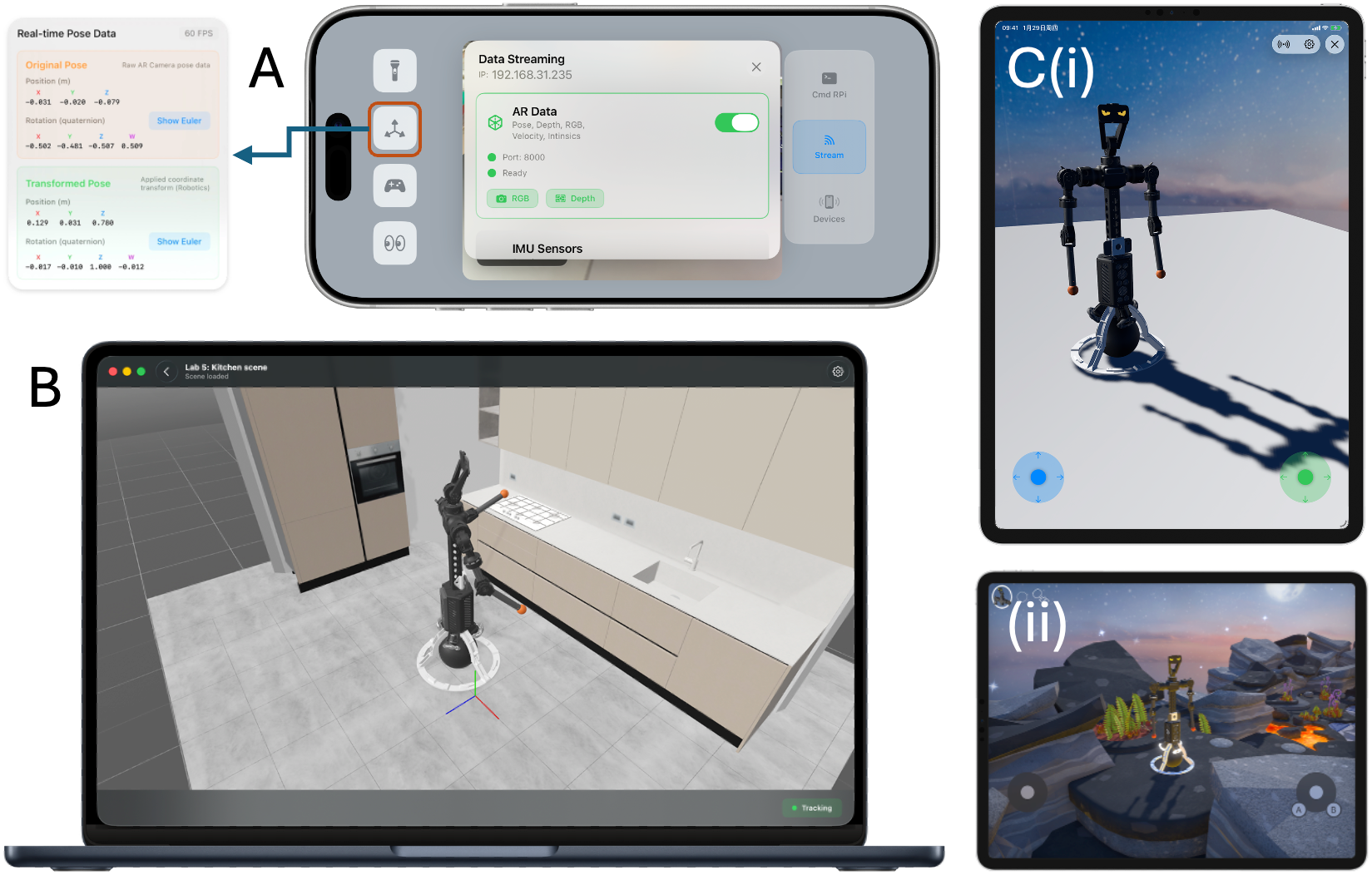}
        \caption{
            Our iOS app ecosystem for multi-modal data streaming and sim-like interactive control.
        }
        \label{fig:iOSApp_Visualization}
    \end{figure}

    We also developed a series of intuitive user interfaces to facilitate the visualization of human-robot interaction. Shown in Fig. \ref{fig:iOSApp_Visualization}A is the data streaming interface when using an iPhone Pro as a robot's sensing suite, which is not limited to asRoBallet but generally applicable by attaching an iPhone to the robot. Fig. \ref{fig:iOSApp_Visualization}B is the MacOS version of our app, enabling users to load OpenUSD models to create a custom simulation environment, such as those in the BEHAVIOUR Challenge \cite{Li2024Behavior1K}. Fig. \ref{fig:iOSApp_Visualization}C is the iPad version of our app, showcasing a game-like visualization where users can either play with a virtual asRoBallet or load asRoBallet into a game for entertainment, where the robot model can be customized.

\section{Results, Demonstrations, and Discussions}
\label{sec:Results}

\subsection{Benchmark Whole-Body Performance of asRoBallet}

    Following the survey in \cite{Jespersen2019Kugle}, which identifies LQR and PI/PD controllers as among the most widely used strategies for ball-balancing robots, we adopt a fixed-gain LQR controller as a representative classical baseline. The LQR was designed and tuned around the nominal upright configuration and achieved stable performance in the nominal setting. We then compare it with the proposed RL policy on velocity-tracking and station-keeping tasks under four levels of increasing difficulty. These levels progressively violate the assumptions of the nominal linear model, including unmodeled changes in the center of mass due to random arm configurations, in the most difficult setting (a robustness stress test outside the fixed-gain LQR design envelope).
    \begin{enumerate}
        \item Fixed default arms with all joints fixed at zero.
        \item Fixed default arms with torsional friction between the ball and the floor, randomly sampled from [0.01, 0.5]. The LQR controller relies on the common assumption that sufficiently large torsional friction prevents rotation about the vertical axis. However, this assumption is occasionally violated in practice, particularly on low-friction surfaces \cite{Jespersen2019Kugle}.
        \item Fixed default arms with random initial body tilt within 5 degrees.
        \item Random arm configurations representing the most challenging setting with significant variations in system inertia and coupling effects.
    \end{enumerate}
    For each configuration, we report the success rate, mean absolute error (MAE), and standard deviation (STD) of the velocity-tracking error, as well as the success rate, steady-state range, and residual speed for station-keeping; the results are summarized in Table \ref{tab:Results_Benchmark}.

    \begin{table*}
        \centering
        \caption{Comparison between LQR controller and RL policy simulation with different task complexities.}
        \label{tab:Results_Benchmark}
        \begin{tabular}{llcccccc}
        \hline
                             &                                   & \multicolumn{3}{c}{Velocity Tracking} & \multicolumn{3}{c}{Station Keeping} \\ \hline
        Methods              & Arm Configuration                 
                             & \multicolumn{1}{c}{\begin{tabular}[c]{@{}c@{}}Success\\ rate\end{tabular}} 
                             & \multicolumn{1}{c}{\begin{tabular}[c]{@{}c@{}}Tracking MAE\\ (cm/s, $10^{-2}$rad/s)\end{tabular}} 
                             & STD 
                             & \multicolumn{1}{c}{\begin{tabular}[c]{@{}c@{}}Success\\ rate\end{tabular}} 
                             & \multicolumn{1}{c}{\begin{tabular}[c]{@{}c@{}}Range\\ (cm)\end{tabular}} 
                             & \multicolumn{1}{c}{\begin{tabular}[c]{@{}c@{}}Speed\\ (cm/s)\end{tabular}} \\ \hline
    
        \multirow{4}{*}{LQR} & Fixed home position $P_0$         
                             & 100 & 7.7 6.0 10.5   & 9.5 9.8 12.7   & 100 & 8.7 & 3.1 \\ \cline{2-8} 
                             & Fixed $P_0$ \& random friction    
                             & 96  & 8.0 6.2 10.4   & 9.8 10.1 12.7  & 18  & 8.1 & 2.9 \\ \cline{2-8} 
                             & Fixed $P_0$ \& random orientation 
                             & 56  & 7.5 5.9 10.5   & 9.1 9.6 12.7   & 0   & /   & /   \\ \cline{2-8} 
                             & Random arm pose                   
                             & 86  & 23.1 23.0 10.8 & 28.7 27.7 12.1 & 6   & 8.4 & 3.3 \\ \hline
    
        \multirow{4}{*}{RL}  & Fixed home position $P_0$         
                             & 100 & 4.0 3.8 1.9    & 7.1 7.4 2.4    & 100 & 4.5 & 3.0 \\ \cline{2-8} 
                             & Fixed $P_0$ \& random friction    
                             & 92  & 3.8 4.0 1.9    & 6.9 7.4 2.4    & 100 & 4.5 & 2.9 \\ \cline{2-8} 
                             & Fixed $P_0$ \& random orientation 
                             & 100 & 4.1 3.8 1.9    & 7.2 7.0 2.4    & 100 & 4.3 & 2.4 \\ \cline{2-8} 
                             & Random arm pose                   
                             & 100 & 6.6 6.2 1.7    & 8.8 8.9 2.1    & 100 & 5.0 & 3.1 \\ \hline
    
        \end{tabular}
    \end{table*}

    \begin{table*}[htbp]
        \centering
        \caption{Ablation study of RL policy in simulation.}
        \label{tab:Results_Ablation}
        \begin{tabular}{lcccccc}
        \hline
                                             & \multicolumn{3}{c}{Velocity Tracking} & \multicolumn{3}{c}{Station Keeping} \\ \hline
        Method                               & \multicolumn{1}{c}{\begin{tabular}[c]{@{}c@{}}Success\\ rate\end{tabular}} 
                                             & \multicolumn{1}{c}{\begin{tabular}[c]{@{}c@{}}Tracking MAE\\ (cm/s, $10^{-2}$rad/s)\end{tabular}} 
                                             & \multicolumn{1}{c}{\begin{tabular}[c]{@{}c@{}}STD\\ \end{tabular}} 
                                             & \multicolumn{1}{c}{\begin{tabular}[c]{@{}c@{}}Success\\ rate\end{tabular}} 
                                             & \multicolumn{1}{c}{\begin{tabular}[c]{@{}c@{}}Range\\ (cm)\end{tabular}} 
                                             & \multicolumn{1}{c}{\begin{tabular}[c]{@{}c@{}}Speed\\ (cm/s)\end{tabular}} \\ \hline
        \multicolumn{7}{l}{\textbf{Ablation on Obs/Reward}} \\ \hline
        arm-obs                    & 92 & 12.0 17.1 2.9 & 15.0 20.0 3.6 & 98 & 8.5 & 5.0 \\ \hline
        track-sigma-0.1            & 96 & 11.5 21.1 1.7 & 13.8 23.8 1.9 & /  & /   & /   \\ \hline
        tanh-reward                & 96 & 6.6 7.8 1.9   & 8.5 10.1 2.3  & \textbf{100} & \textbf{4.5} & \textbf{2.6} \\ \hline
        exp-reward                 & \textbf{96} & \textbf{5.8 6.3 1.7} & \textbf{8.4 8.9 2.1} & 96 & 7.7 & 4.4 \\ \hline
        \multicolumn{7}{l}{\textbf{Ablation on Sim-to-Real}} \\ \hline
        w/o-imu-noise              & 78 & 5.6 6.8 1.7 & 8.3 9.7 2.1 & 96 & 5.1 & 2.8 \\ \hline
        w/o-friction-randomization & 82 & 7.2 7.1 2.0 & 8.6 9.4 2.3 & 98 & 5.9 & 3.4 \\ \hline
        asRoBallet                 & \textbf{96} & \textbf{5.8 6.3 1.7} & \textbf{8.4 8.9 2.1} & \textbf{100} & \textbf{4.5} & \textbf{2.6} \\ \hline
        \end{tabular}%
    \end{table*}

    Under nominal conditions (fixed-arm configuration), both controllers achieve a 100\% success rate. However, the RL policy consistently achieves substantially lower tracking errors, with an MAE of 4.0 cm/s compared with 7.7 cm/s for LQR, and a lower variance. As task difficulty increases, the performance gap becomes more pronounced. In the presence of random torsional friction, the LQR success rate drops to 96\%, whereas RL maintains high accuracy with a lower MAE and STD. When random initial orientations are introduced, the LQR controller degrades significantly, achieving only a 56\% success rate, while the RL controller remains robust with a 100\% success rate. In the most challenging scenario with random arm configurations, the LQR controller exhibits large tracking errors (MAE exceeding 23 cm/s) and high variability, whereas the RL controller preserves stable performance with a 100\% success rate and comparatively low error.

    A similar trend is observed for station keeping. While both controllers perform well under nominal conditions, the LQR controller degrades rapidly when assumptions are violated. In scenarios involving random friction or orientation perturbations, the LQR success rate drops sharply (to 18\% and 0\%, respectively) and, in some cases, fails entirely. In contrast, the RL controller consistently achieves a 100\% success rate across all configurations, maintaining a smaller steady-state position range and lower residual velocity. Notably, even under random arm configurations, RL preserves stable station-keeping behavior, whereas LQR performance degrades substantially.

\subsection{Ablation Study}

    We further conducted ablation studies to isolate the contribution of key design choices in the RL policy, covering: (i) observation/reward shaping, and (ii) domain randomization for sim-to-real robustness. All ablated variants were trained under the same protocol as the full RL method (denoted asRoBallet) and evaluated on the same velocity-tracking and station-keeping benchmarks reported in Table \ref{tab:Results_Ablation}. All evaluation tasks were executed after removing the virtual ball joint between the torso and the ball. Each ablation task was repeated 50 times, with randomizations applied to the episode reset.

    \textbf{Ablation on Observation/Reward Design.} Arm-obs: replaces the body-centric state representation by directly using the 8 arm joints' observed positions, instead of the CoM–related features used in the full design. This ablation test assesses whether the policy benefits more from a task-relevant global representation (CoM) than from high-dimensional arm-configuration signals. Tanh-reward: uses a tanh-shaped tracking reward, which saturates for large errors and therefore reduces gradient sensitivity to outliers. Exp-reward: uses an exponential tracking reward, providing stronger shaping near the target region. Track-sigma-0.1: changes the exponential reward sharpness by setting $\sigma=0.1$.

    \textbf{Ablation on Domain Randomization.} w/o-imu-noise: removes injected IMU noise during training, to test whether sensor perturbations are necessary for policy robustness. w/o-friction-randomization: removes friction randomization (e.g., contact/torsional friction variations) during training, to test whether contact uncertainty is a key driver of generalization.

    Replacing the CoM-related state with raw arm joint positions substantially degrades performance (velocity-tracking success drops to 92\% with a much larger MAE/STD, and station-keeping shows greater drift), indicating that a compact, task-relevant global representation is critical. Reward shaping mainly affects the tracking–holding trade-off: the exponential tracking reward achieves the best tracking accuracy (MAE 5.8/6.3/1.7 cm/s with low STD, matching the full model), while the tanh-shaped reward slightly reduces tracking accuracy but yields the best station keeping (100\% success, 4.5 cm range, 2.6 cm/s residual speed). In contrast, an overly sharp exponential reward ($\sigma=0.1$) harms tracking, suggesting brittle optimization when deviations are penalized too aggressively. Removing domain randomization markedly reduces robustness, especially without IMU noise (tracking success 78\%) and without friction randomization (tracking success 82\%), highlighting that realistic sensing/contact perturbations are essential for reliable generalization.

    \begin{figure}[h]
        \centering
        \includegraphics[width=1\linewidth]{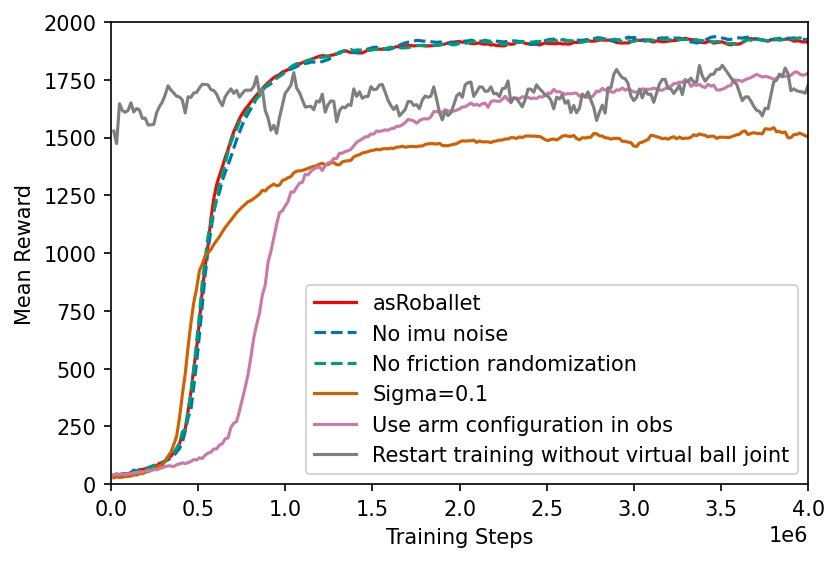}
        \caption{Velocity-tracking learning curves (mean episode reward) for the proposed RL policy and ablations.}
        \label{fig:Results_Ablation}
    \end{figure}

    Figure \ref{fig:Results_Ablation} shows the learning curves (mean episode reward) for the velocity-tracking task under the proposed RL design and several ablations. Although removing IMU noise or applying friction randomization during training yields curves that are nearly indistinguishable from those in the proposed setting in terms of final reward and convergence speed, these modifications are important for achieving robust success rates at evaluation time. 

    In addition, we examined whether the policy could be further fine-tuned after removing the virtual ball joint. Specifically, we restarted training from the checkpoint at $4.0\times10^6$ steps, where a stable policy had already been obtained, and continued training without the virtual ball joint (gray curve, ``restart without ball joint''). This setting produced highly variable rewards, showed no clear monotonic improvement, and remained consistently below the performance ceiling achieved by the proposed method.
    
    These results suggest that the low-damping virtual ball joint serves as a training-time stabilizer that facilitates effective exploration. Due to the severe underactuation of the ballbot system, unstructured early exploration can easily drive the system away from valid wheel-ball contact states, thereby greatly reducing the probability of collecting informative transitions and weakening the learning signal. The two empirical observations, namely i) training from scratch without the virtual ball joint and ii) removing the joint after initial convergence, both indicate that in our current setup the virtual joint stabilizes the learning process, while its removal destabilizes further training. Importantly, the virtual joint is used only during training; all evaluations and hardware deployment are conducted without it. We therefore regard the joint as a training curriculum/stabilizer rather than a deployment crutch.
    
\subsection{Sim2Real Transfer of Underactuated Spherical Dynamics}

\begin{figure}[htbp]
        \centering
        \includegraphics[width=1\linewidth]{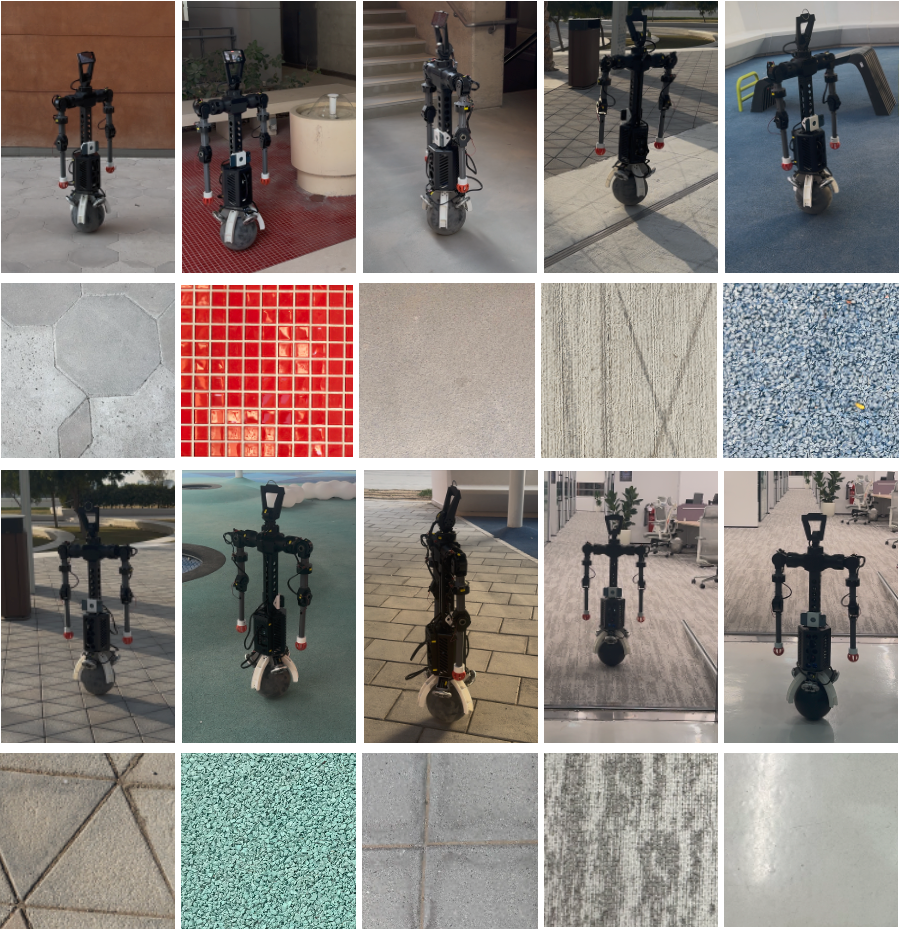}
        \caption{Successful Sim2Real deployment of asRoBallet on various indoor \& outdoor floor textures and lighting conditions.}
        \label{fig:Results_RealWorld}
    \end{figure}

    We achieved zero-shot Sim2Real transfer of our RL policy during hardware deployment across various floor textures under indoor and outdoor lighting conditions, as shown in Fig. \ref{fig:Results_RealWorld}. asRoBallet moves smoothly in these common floor conditions we found around the campus, and can even pass through slight obstacles or move across different floor textures. One example is the last two images of Fig. \ref{fig:Results_RealWorld}, where asRoBallet successfully moved down a 5-degree slope on office carpet, overcame a slight aluminum bump on the door frame, and then transitioned to a hard, anti-slip floor. Please refer to our supplementary video for further demonstration. 

    The testing process is not without failure, especially when there is a sudden increase in floor height. When it falls, the 3D-printed neck connector appears to be the most vulnerable component. However, the whole system exhibited exceptional robustness and survived several strong impacts, remaining operational almost immediately after the fall. It should be noted that in the tests shown in Fig. \ref{fig:Results_RealWorld}, we \textit{purposely} removed the protective ring around the ball in Fig. \ref{fig:Design_asRoBallet} to stress test asRoBallet's performance and our proposed learning method.

\begin{figure}[htbp]
        \centering
        \includegraphics[width=1\linewidth]{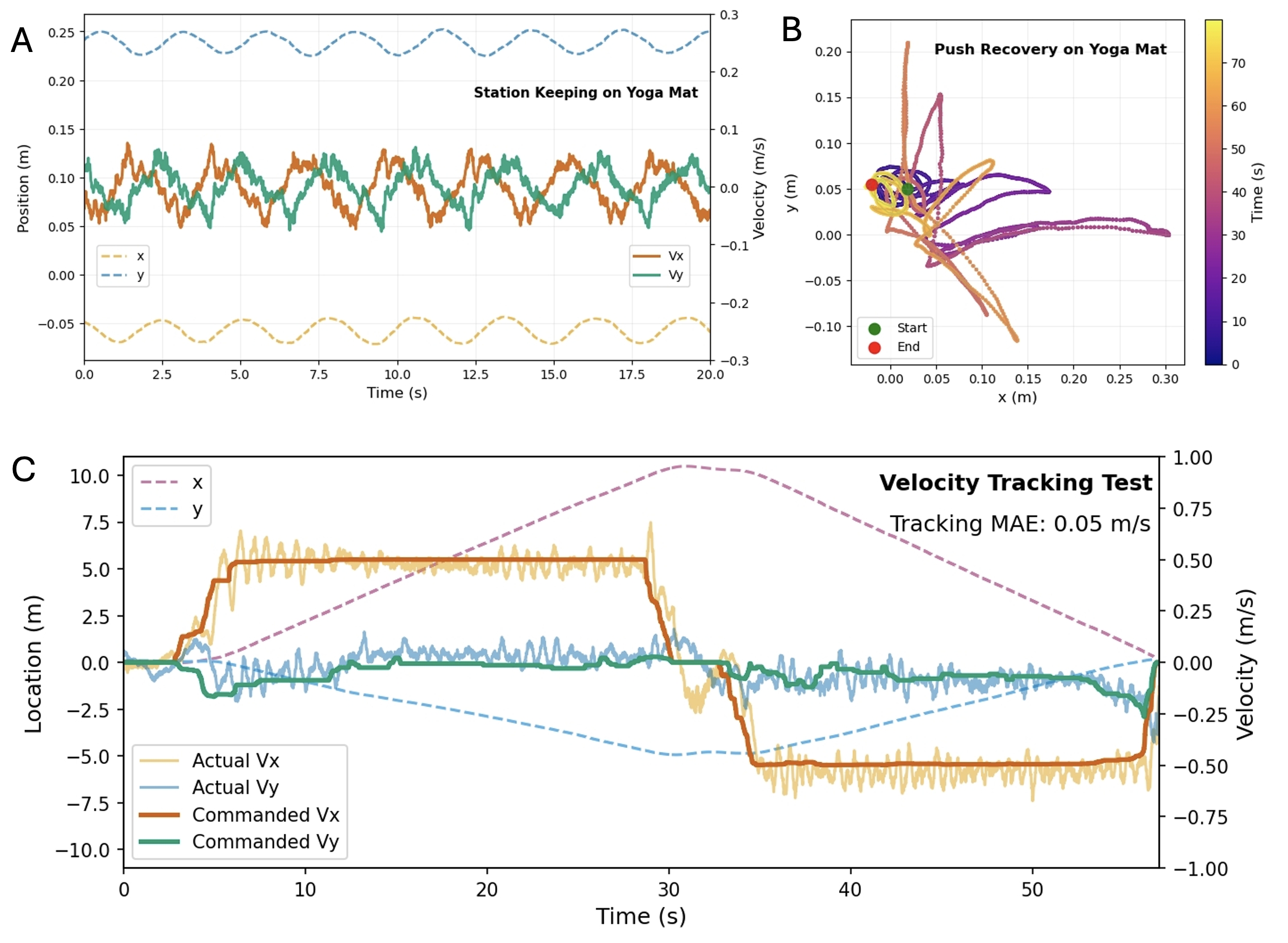}
        \caption{Performance of real robot experiments.}
        \label{fig:Results_RealWorldMetrics}
    \end{figure}

    We evaluated the stationary-keeping policy on asRoBallet across two floor surfaces: ceramic tiles and a yoga mat (Fig. \ref{fig:Results_RealWorldMetrics}A), where the position variation remained within 5 cm and 3 cm, respectively, and the average residual speeds were 5.1 cm/s and 4.4 cm/s, respectively. We further tested the push recovery on the yoga mat. asRoBallet managed to return to the starting point after being pushed from different directions as far as 0.3 meters away in 7 trials (Fig. \ref{fig:Results_RealWorldMetrics}B). Hardware results on velocity tracking are shown in Fig. \ref{fig:Results_RealWorldMetrics}C, where asRoBallet is controlled by a human operator via a joystick on the common office floor carpet and achieved a mean absolute error of 0.05 m/s.

\section{Conclusion, Limitations, and Future Work} 
\label{sec:conclusion}

    This work presents \textit{asRoBallet}, a holistic system that overcomes the barriers to deploying reinforcement learning on underactuated spherical robots. By closing the \textit{Reality Gap} inherent in the complex tribology of wheel-ball-floor interactions, we, to the best of our knowledge, achieved the \textit{first} end-to-end RL locomotion policy deployed on a humanoid ballbot hardware platform. We validated \textit{subtractive reconfiguration} as a viable design strategy, demonstrating that high-performance humanoid research platforms can be constructed cost-effectively by repurposing the proprioceptive, computational, and perceptual cores of an overconstrained quadruped. We showed that the primary bottleneck for sim2real ballbot RL is not the learning algorithm itself, but the fidelity of the contact model. By explicitly simulating the discrete mechanics of the roller and the parasitic vibrations of ETH-type omni-wheels in MuJoCo, our \textit{Friction-Aware} framework outperforms classical LQR baselines in robustness to unknown surface textures and calibration errors. We bridged the usability gap by integrating consumer-grade mobile computing directly into the control loop. The \textit{iOS app} ecosystem demonstrates that pervasive devices (i.e., iPhones) can serve as low-latency, high-bandwidth interfaces for whole-body telepresence, democratizing access to advanced humanoid control.
    
    Despite these successes, several \textit{limitations} remain in the current implementation. As noted in our ablation studies, the policy fails to converge without a virtual ball joint between the torso and the ball during training. This dependence suggests that the reward landscape for spherical balancing is exceptionally sparse, requiring artificial constraints to guide the agent through the initial exploration phase. While our policy manages station-keeping effectively, the Coulomb friction creates a \textit{dead zone} that necessitates continuous micro-oscillations for stabilization. While our RL policy learns to manage this better than LQR does, it still yields residual jitter inherent to the current drivetrain configuration.
    
    Future work will extend asRoBallet from robust locomotion to mobile manipulation by incorporating the currently passive upper-body joints into the active control loop. We plan to train unified policies over the spherical base and 4-DoF arms for dynamic tasks such as door opening and object transport, where the robot must exploit body motion to generate manipulation forces. We also aim to reduce the reality gap by learning vision-to-torque policies from raw iPhone inputs, enabling implicit adaptation to surface properties such as carpet or tile. Finally, using the peer-to-peer Wi-Fi Aware capabilities of the iOS ecosystem, we will explore multi-agent collaborative balancing for large-object transport and navigation in crowded social spaces.

\section*{Acknowledgments}

    This work was partially supported by Shenzhen AncoraSpring Robotics Technology Co., Ltd., National Natural Science Foundation of China (62473189), and Guangdong Basic and Applied Basic Research Foundation under Grant 2025A1515010424.

\bibliographystyle{unsrtnat}
\bibliography{References}
\end{document}